\documentclass[runningheads]{llncs}
\usepackage{graphicx}

\usepackage{tikz}
\usepackage{comment}
\usepackage{amsmath,amssymb}

\usepackage{graphicx}
\usepackage{amsmath}
\usepackage{amssymb}
\usepackage{booktabs}
\usepackage{subfigure}
\usepackage{multirow}
\usepackage{thmtools}
\usepackage{enumitem}
\usepackage{caption}
\usepackage{algorithm}     
\usepackage{algorithmicx}
\usepackage[noend]{algpseudocode} 
\usepackage{booktabs}
\usepackage[export]{adjustbox}

\usepackage[pagebackref,breaklinks]{hyperref}

\usepackage[capitalize]{cleveref}
\crefname{section}{Sec.}{Secs.}
\Crefname{section}{Section}{Sections}
\Crefname{table}{Table}{Tables}
\crefname{table}{Tab.}{Tabs.}

\newcommand{\decorate}[1]{\texttt{#1}}
\newcommand{\Triplet}[3]{$\langle\decorate{#1}, \decorate{#2}, \decorate{#3}\rangle$}

\newcommand{\etal}{\textit{et.al.}}

\begin{document}
\mainmatter

\title{Chairs Can be Stood on: Overcoming Object Bias in
Human-Object Interaction Detection} 

\author{Guangzhi Wang\inst{1} \and
Yangyang Guo\inst{2}\thanks{corresponding author} \and
Yongkang Wong\inst{2} \and 
Mohan Kankanhalli\inst{2}}
\authorrunning{Wang et al.}
\titlerunning{Overcomming Object-Bias in Human-Object Interaction Detection}

\institute{Institute of Data Science, National University of Singapore \and
School of Computing, National University of Singapore \\
\email{guangzhi.wang@u.nus.edu guoyang.eric@gmail.com \\
yongkang.wong@nus.edu.sg  mohan@comp.nus.edu.sg}
}

\maketitle

\begin{abstract}
Detecting Human-Object Interaction (HOI) in images is an important step towards high-level visual comprehension.
Existing work often shed light on improving either human and object detection, or interaction recognition.
However, due to the limitation of datasets, these methods tend to fit well on frequent interactions 
conditioned on the detected objects, yet largely ignoring the rare ones,
which is referred to as the \textbf{object bias problem} in this paper.
In this work,
we for the first time, uncover the problem from two aspects: unbalanced interaction distribution and biased model learning.
To overcome the object bias problem, we propose a novel plug-and-play  Object-wise Debiasing Memory (ODM) method for re-balancing the distribution of interactions under detected objects.
Equipped with carefully designed read and write strategies, the proposed ODM allows rare interaction instances to be more frequently sampled for training, thereby alleviating the object bias induced by the unbalanced interaction distribution. 
We apply this method to three advanced baselines and conduct experiments on the HICO-DET and HOI-COCO datasets.
To quantitatively study the object bias problem, 
we advocate a new protocol for evaluating model performance.
As demonstrated in the experimental results, our method brings consistent and significant improvements over baselines, especially on rare interactions under each object.
In addition, when evaluating under the conventional standard setting, our method achieves new state-of-the-art on the two benchmarks.
\end{abstract}

\section{Introduction} \label{sec:intro}

Benefiting from the advancement of visual detection systems,
Human-Object Interaction (HOI) detection has drawn increasing research interests in recent years.
It requires detecting both humans and objects in a given image, based on which the interactions (often expressed as verb phrases) should also be correctly recognized.
HOI detection is of vital importance to human-centric visual understanding and also benefits other high-level vision tasks, such as image captioning~\cite{li2017scene} and visual question answering~\cite{antol2015vqa,guo2019quantifying}.

\begin{figure}[t]
    \centering
    \includegraphics[width=0.98\columnwidth, trim=0 0 40 0, clip]{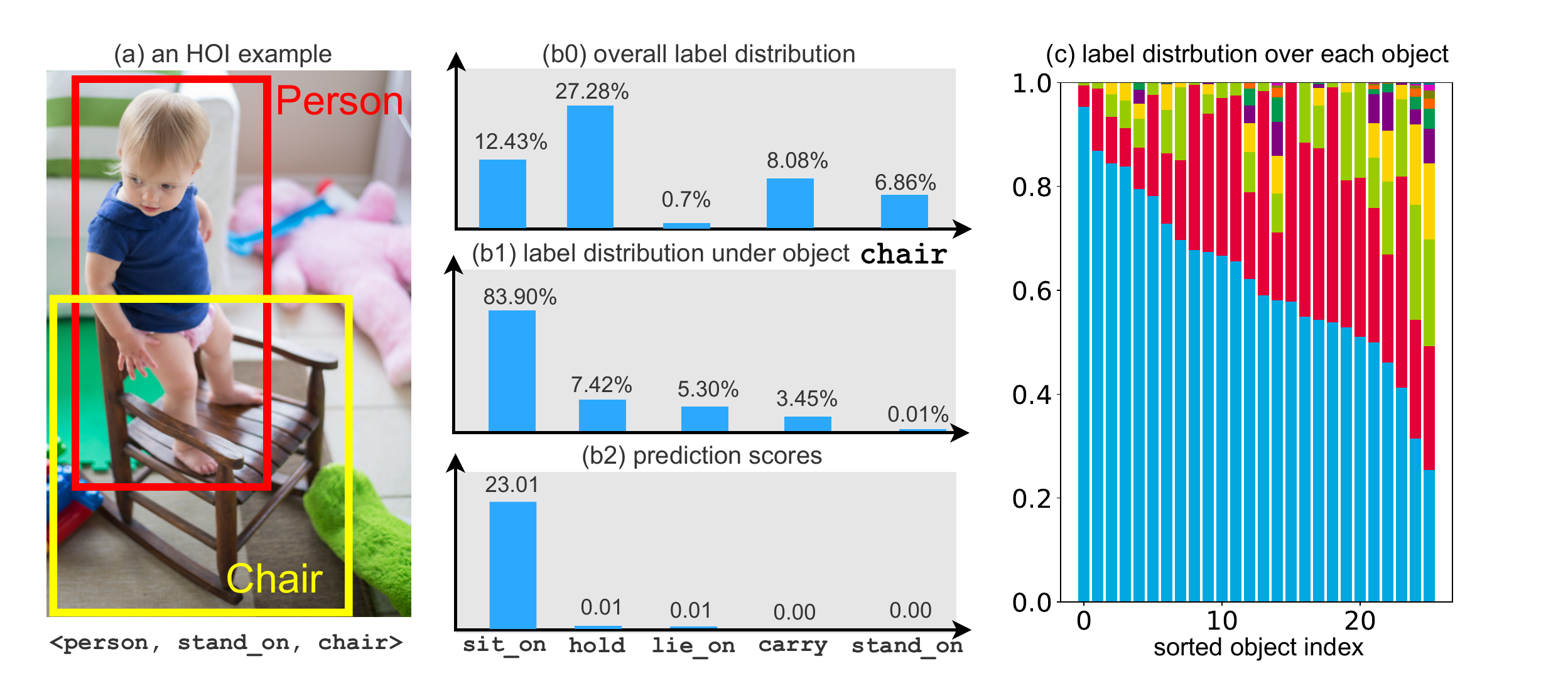}
    \caption{\label{fig:teaser}
    An illustration of the object bias problem.
    Given the detected human-object pair in (a), the model~\cite{zhang2021spatially_SCG_ICCV21} prediction (b2) is highly biased towards the object-conditional label distribution (b1), instead of the overall long-tail distribution in the training set (b0).
    As a result, the model predicts a more frequent verb \decorate{sit\_on} for the object \decorate{chair}, leaving the true label \decorate{stand\_on} ignored.
    (c) Label distribution from 25 randomly selected objects.
    It can be seen that most objects are dominated by one interaction (colored in blue).
    }
    \vspace{-20pt}
\end{figure}

Existing HOI detection efforts can be mainly categorized into two groups: two-stage and one-stage methods.
Specifically, methods in the first group often leverage an off-the-shelf detector (\textit{e.g.},~Faster R-CNN~\cite{ren2015fasterrcnn}) to initially detect the regions of humans and objects.
The succeeding stage of interaction recognition can be enhanced with human part/pose understanding~\cite{gupta2019nofrills_ICCV19,wan2019poseaware,li2020djrn_CVPR20,dong2021visualpartsum}, graph-based message passing between humans and objects~\cite{qi2018learning_GPNN_ECCV18,zhou2019relation_RPNN_ICCV19,gao2020drg_ECCV20,wang2019contextualattention_ICCV19,ulutan2020vsgnet_CVPR20,zhang2021spatially_SCG_ICCV21} or finer label space construction~\cite{kim2021acp++_TIP,zhong2020polysemy_ECCV20}.
Some studies also exploit cross-dataset knowledge such as human-object interactiveness~\cite{li2019transferable_TIN_CVPR19,li2021transferable_TPAMI,xu2019interact_iHOI_TMM}, cross-dataset objects~\cite{hou2021affordance_ATL_CVPR21} and word embeddings~\cite{xu2019learning_knowledge_CVPR19} to improve interaction recognition.
Nonetheless, these approaches are often limited by deficiencies like inferior proposal generation or heavy inference overhead.
To address these problems, one-stage methods often resort to performing detection and interaction classification within a single stage.
Early studies treat HOI detection as a \Triplet{human}{object}{interaction} point detection and matching~\cite{liao2020ppdm_CVPR20,wang2020learning_IPNet_CVPR20,zhong2021glance_GGNet_CVPR21} task. 
Recent approaches employ the Transformer-based detector~\cite{carion2020detr} to aggregate contextual information and detect interaction in an anchor-free manner~\cite{tamura2021qpic_CVPR21,zou2021end_HOITransformer_CVPR21,chen2021reformulating_ASNet_CVPR21,kim2021hotr_CVPR21,zhang2021mining_CDN_NIPS21}.
Nevertheless, increased training time is often encountered by this group of approaches.

Although existing methods have made progress over benchmarks, we observe one pervasive shortcoming that prevents them from further advancement.
That is, the interaction prediction is strongly related to the detected object.
Fig.~\ref{fig:teaser} shows that given the detected object \decorate{chair} in (a), the model predicts (b2) the wrong verb \decorate{sit\_on} with a very high confidence, rather than yields the true action - \decorate{stand\_on}. 
Previous studies~\cite{zhang2021mining_CDN_NIPS21,hou2020visual_VCL_ECCV20,hou2021affordance_ATL_CVPR21} mostly perceive this phenomenon as the outcome of learning from the long-tail label distribution from the overall training set.
Nevertheless, as we step further into this problem, we find it deviates a lot from the intuition of those methods.
In particular, as shown in Fig.~\ref{fig:teaser} (b0), the label \decorate{hold} dominates the training set and is twice frequent than \decorate{sit\_on}.
Out of expectation, the prediction score (Fig.~\ref{fig:teaser} (b2)) for \decorate{hold} is only 0.01, which is 2,000 times smaller than that of \decorate{sit\_on}.
This observation brings our concern - is the wrong prediction really because of the long-tail label distribution in the overall training set? 
With this concern, we shift our focus to the interaction distribution under the detected object (Fig.\ref{fig:teaser} (b1)), and discover a strong bias between the object and its conditional interaction distribution.
Specifically, the model prediction conforms more with such object-induced bias, rather than the bias caused by the overall long-tail label distribution.
In view of this, we can infer that during training,
the object-induced bias drives the model to fit well on frequent interactions under each object, while overlooking the rare ones.
However, rare classes are often more informative than non-rare ones~\cite{NEURIPS2020_longtail_momentum,wu2020dbloss_ECCV20}.
Simply ignoring them undermines the model's representation ability, resulting in poor generalization and limited real-world applicability.
Nonetheless, to the best of our knowledge, this bias problem has not been explored in the existing literature.
As most objects struggle with the biased interaction distribution (Fig.~\ref{fig:teaser} (c)), we therefore humbly suggest this problem to the community, and name it as the \textit{object bias problem} in this work.

As a matter of fact, dealing with this problem is non-trivial due to the inherent distribution imbalance in existing benchmarks.
However, building a balanced dataset is time and labor intensive.
One alternative solution is to feed the model with balanced samples during training, which has been extensively proved effective in previous studies~\cite{shen2016relaybp,wu2020dbloss_ECCV20,li2019repair}.
Yet, directly applying these methods to HOI detection is sub-optimal, as the object bias problem is actually induced by the class imbalance \textit{under each object}, rather than that of the overall training set.
To this end, we propose a novel Object-wise Debiasing Memory (ODM) module to achieve object-conditional class balancing.
The proposed ODM is implemented with an object-indexed memory, upon which read and write strategies are designed to support the retrieval and storage of HOI features and labels.
For memory reading, we take the label of each interactive instance as query to retrieve instances from the memory. 
Our read strategy assures that rare class instances are more frequently sampled, 
leading to a more balanced label distribution within the batch for training.
On the other hand,
the writing strategy is devised to store rare class instances with higher probability.
In this way, the unbalanced interaction distribution under each object is mitigated, thus reducing the influence of the \textit{object bias problem}.

We conduct extensive experiments over two benchmark datasets, namely HICO-DET~\cite{HORCNN_Chao2018WACV} and HOI-COCO~\cite{hou2021affordance_ATL_CVPR21}. 
In addition, we also advocate a new object bias evaluation protocol to quantitatively evaluate the model performance under the object-biased condition.
When equipped with our method, several advanced baselines are evidently shown to overcome the object bias problem, thereby achieving improved performance.

To summarize, our contributions are three-fold:
\begin{itemize}
    \item We systematically study the object bias problem in the HOI detection task. 
    To the best of our knowledge, we are the first to recognize and address this problem in the HOI literature.
    \item To alleviate the object bias problem, we propose a novel ODM module to facilitate the learning of a balanced classifier.
    The proposed ODM is model-agnostic and applicable to both one-stage and two-stage methods.
    \item We conduct extensive experiments on benchmark datasets, namely HICO-DET~\cite{HORCNN_Chao2018WACV} and HOI-COCO~\cite{hou2021affordance_ATL_CVPR21}.
    When applying our method to several baselines, significant performance improvements, especially on rare interactions under each object, can be observed.
    As a side product, we achieve new state-of-the-art performance on the two datasets\footnote{Code available: \url{https://github.com/daoyuan98/ODM}.}.
\end{itemize}

\section{Related Work} \label{sec:related}

\subsection{Human-Object Interaction Detection}
HOI detection~\cite{HORCNN_Chao2018WACV} is challenging since it requires both precise detection and complex interaction reasoning capabilities.
Existing methods have achieved some progress and mainly fall into two groups: two-stage and one-stage methods.

Two-stage methods adopt an off-the-shelf detector to perform detection, followed by an interaction prediction model over each human-object pair
\cite{HORCNN_Chao2018WACV,gao2018ican_BMVC18,ulutan2020vsgnet_CVPR20,li2021transferable_TPAMI}.
Previous approaches mostly endeavor to improve visual feature quality for interaction classification.
For example, Qi~\etal~\cite{qi2018learning_GPNN_ECCV18} builds a holistic graph to assist information flow for all humans and objects, and Zhang~\etal~\cite{zhang2021spatially_SCG_ICCV21} devises a bipartite graph utilizing relative spatial relation to promote interaction understanding. 
Besides, compositional models factorize the verb and object classification branches to improve generalization~\cite{li2020hoi_IDN,hou2021affordance_ATL_CVPR21,hou2021detecting_FCL_CVPR21}.
Beyond the visual appearance, more complementary cues are explored for the second stage, such as human pose and parts~\cite{li2020djrn_CVPR20,liu2020amplifying_ECCV20,gupta2019nofrills_ICCV19}, language embeddings~\cite{kim2020detecting_ACP_ECCV,bansal2020functional,xu2019learning_knowledge_CVPR19} and external knowledge~\cite{li2019transferable_TIN_CVPR19,he2021exploiting_SG_HOI_ICCV21}.

One-stage methods perform both detection and interaction classification in an end-to-end manner. 
Besides detecting human and object regions, earlier one stage methods exploit either human-object interaction points~\cite{liao2020ppdm_CVPR20,wang2020learning_IPNet_CVPR20} or their union regions \cite{kim2020uniondet_ECCV20} as interaction clues.
With the success of Transformer\cite{transformer} for object detection~\cite{carion2020detr}, some methods ~\cite{chen2021reformulating_ASNet_CVPR21,zou2021end_HOITransformer_CVPR21,kim2021hotr_CVPR21,tamura2021qpic_CVPR21,zhang2021mining_CDN_NIPS21} present to formulate HOI detection as a set-prediction problem, where the anchor-free detection and attention-based global context aggregation are jointly operated.

Recently, some studies focus on the long-tail distribution problem in HOI detection benchmarks.
For example, ATL~\cite{hou2021affordance_ATL_CVPR21} constructs new HOI instances from external object datasets in an affordance transfer fashion, while FCL~\cite{hou2021detecting_FCL_CVPR21} generates object features to fabricate more training samples.
Besides, CDN~\cite{zhang2021mining_CDN_NIPS21} presents a dynamic re-weighting mechanism to tackle the long-tail problem.
However, they mainly focus on the general long-tail distribution from the whole training set, leaving the \textit{object bias problem} untouched in the literature.

\subsection{Bias Identification and Mitigation}
Previous practices on the bias problem mainly follow an identification then mitigation paradigm.
Pertaining to the bias identification, Zhao~\etal~\cite{zhao2017menlikeshopping} finds that the gender bias contained in datasets can be further amplified by the model trained on them.
Manjunatha~\etal~\cite{manjunatha2019explicit_rulemining} explicitly discovers the bias in Visual Question Answering~\cite{antol2015vqa} via association rule mining, while Guo~\etal~\cite{guo2021loss} alleviates the bias through loss re-scaling. 
Lately, Li and Xu~\cite{LiUICCV2021_discoverunknown_bias} unearths unknown biased attributes of a classifier with generative models.
To mitigate the bias problem, adversarial training~\cite{ganin2015unsupervised} is employed to learn bias irrelevant representations~\cite{zhang2018mitigating}.
Recently, Wang~\etal~\cite{wang2020benchmarkbias} benchmarks previous mitigation methods and presents a combination of domain-conditional models for de-biasing, 
while Choi~\etal~\cite{choi2020fair} tackles the unbalanced distribution with the weak supervision from a small reference dataset. 

\vspace{-1em}
\section{Object Bias Identification}\label{sec:bias_analysis}
\vspace{-1em}
The \textit{object bias problem} in HOI detection refers to predicting interactions based on the unbalanced label distribution under each object. 
In the following, we demonstrate that the \textit{object bias} problem comes from two aspects: 
(1) the conditionally unbalanced label distribution induced by objects and 
(2) the biased model training on the datasets.


\vspace{-1em}
\subsection{Unbalanced Verb Distribution} 
\begin{figure*}[ht]
    \vspace{-20pt}
    \centering
    \includegraphics[width=1.0\textwidth, trim=0 10 10 0, clip]{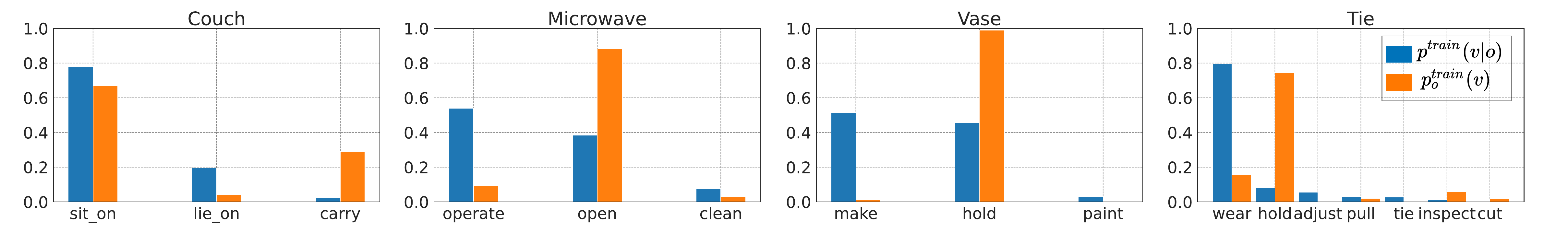}
    \caption{\label{fig:objbias31_dist}
     Comparison between the object-conditional verb distribution $p^{train}(v|o)$ and the overall re-normalized distribution $p^{train}_o(v)$ for four objects in the training set of HICO-DET~\cite{HORCNN_Chao2018WACV}.
    }
    \vspace{-10pt}
\end{figure*}

The objective of HOI detection is to detect and classify \Triplet{human}{verb}{object} triplets, where the most challenging and crucial part is verb classification.\footnote{With the detected object, verb classification is required to recognize the interaction.}
Denote the whole verb set as $\mathcal{V}$,
$\mathcal{V}_o$ for object $o$ represents a subset of all verbs, i.e., $\mathcal{V}_o \in \mathcal{V}$ and $|\mathcal{V}_o| < |\mathcal{V}|$.
We use $p(v|o)$ to represent the verb distribution conditioned on object $o$, and
$p_o(v)$ to signify the global verb distribution involving only verbs for object $o$ in the training set.
The latter is employed to re-normalize the number of verbs associated with $o$, leaving other irrelevant verbs unaffected.
From Fig.~\ref{fig:objbias31_dist}, we can observe that these two distributions are both skewed and actually different from each other.
Besides, a globally frequent class can be a rare one after object conditioning and vice versa.
For example, \decorate{hold} is the most frequent verb from a global view, while the object \decorate{vase} sees verb \decorate{make} most (Fig.~\ref{fig:objbias31_dist}).
It thus brings our question: among these two long-tailed distributions, which one dominates more for the final verb classification?

\vspace{-1.5em}
\subsection{Biased Model Learning}
\begin{figure}
    \vspace{-2em}
    \centering
    \includegraphics[width=1.0\textwidth]{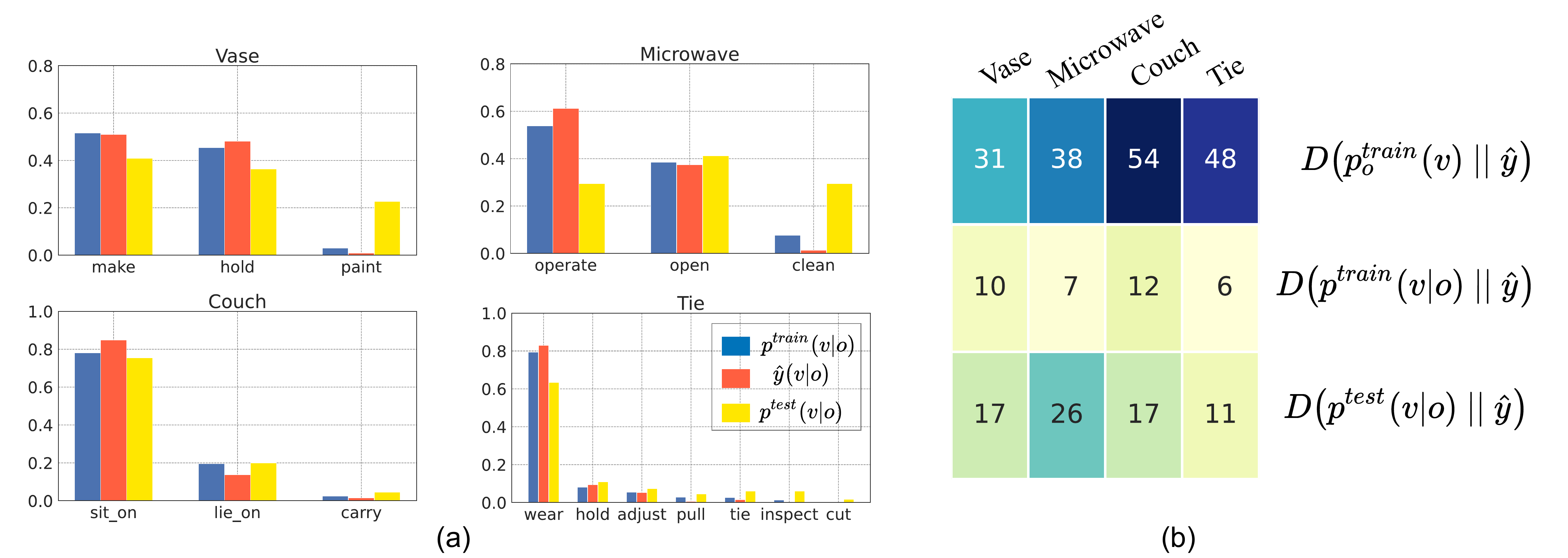}
    \caption{\label{fig:model_output}
    (a) The model output analysis of SCG~\cite{zhang2021spatially_SCG_ICCV21} on HICO-DET~\cite{HORCNN_Chao2018WACV} test set with four objects.
    We show the difference between conditional training distribution $p^{train}(v|o)$, averaged model output $\hat{y}(v|o)$ and the ground-truth conditional verb distribution $p^{test}(v|o)$.
    (b) The Jensen-Shannon divergence~\cite{lin1991jsd} is utilized to compute the distribution distance.
     Note that the values are increased 100x for better illustration.}
    \vspace{-3em}
\end{figure}

To delineate the second aspect, we exemplarily study the behavior of the state-of-the-art model SCG~\cite{zhang2021spatially_SCG_ICCV21} on the HICO-DET dataset.
Denote $\hat{y}(v|o)$ as the averaged verb score output by SCG conditioned on object $o$, and $p^{test}(v|o)$ as the counterpart of $p^{train}(v|o)$ on the test set.
We compare them in Fig.~\ref{fig:model_output} (a)
and observe that $\hat{y}(v|o)$ pays less attention to conditionally rare classes in the training set (e.g., \decorate{paint} for \decorate{vase}, \decorate{clean} for \decorate{microwave}).
In contrast, conditionally frequent classes 
(e.g., \decorate{operate} for \decorate{microwave}, \decorate{wear} for  \decorate{tie} and \decorate{sit on} for  \decorate{couch}) gain higher scores regardless of their prediction correctness.
To quantify how much \textit{bias} the model has learned, we compute the Jensen-Shannon Divergence~\cite{lin1991jsd} between $\hat{y}(v|o)$ and $p_o^{train}(v)$, $p^{train}(v|o)$, $p^{test}(v|o)$ and visualize them in Fig.~\ref{fig:model_output} (b).
We can see that $\hat{y}(v|o)$ is closer to $p^{train}(v|o)$ than $p^{train}_o(v)$, indicating the model leans towards the object-conditional statistics, rather than the overall label distribution in the training set.
Besides, $\hat{y}(v|o)$ is even more similar to $p^{train}(v|o)$ than the ground-truth distribution $p^{test}(v|o)$.
This implies, if we can counteract the learning of the $\textit{object bias}$, there is a large potential of performance improvements with existing methods.


\subsection{Comparison with Other Biases}


\textbf{Long-tail in HOI Detection} We notice that some prior efforts~\cite{zhang2021mining_CDN_NIPS21,hou2021affordance_ATL_CVPR21,hou2020visual_VCL_ECCV20} have studied the long-tail problem in HOI detection.
Nevertheless, the \textit{object bias problem} presented in this paper is intrinsically and technically different from the long-tail one.
On the one hand, the object-conditional distribution can be distinct from the overall long-tail distribution. 
For example, \decorate{hold} is the most frequent verb across the whole dataset but less frequent in some objects (Fig.~\ref{fig:objbias31_dist}).
On the other hand, the model prediction tends to conform with the object-conditional label distribution, rather than the overall one.
Combing these two sides, we thus introduce the \textit{object bias problem} to the community and expect more insightful findings along this line.

\noindent\textbf{Bias in Scene Graph Generation (SGG)} It is worth noting that the bias problem in the sister task - SGG, is also different from the \textit{object bias problem}.
In fact, mainstream studies in SGG debiasing ~\cite{yan2020pcpl,chiou2021recovering,Tang_2020_CVPR_unbiased_SGG} mainly focus on the overall class imbalance, which is essentially same as the long-tail problem in HOI detection.
The most relevant work to ours is ~\cite{zellers2018neuralmotif}.
It leverages the most frequent predicate under \decorate{subject}-\decorate{object} pair for relation prediction, which is shown to be a strong baseline on benchmark dataset.
However, there are two key differences between ~\cite{zellers2018neuralmotif} and our work:
1)~\cite{zellers2018neuralmotif} focuses on relational bias from the data's perspective only,
while we provide a comprehensive study across the aspects of the dataset, model behavior and evaluation protocol.
2)~\cite{zellers2018neuralmotif} leverages the training set statistics to conduct prediction. 
However, when deploying the method to another dataset or other out-of-distribution settings, degraded performance is expected, as it severely overfits specific training set~\cite{chen2019knowledge,Tang_2020_CVPR_unbiased_SGG}.
By contrast, we design a novel debiasing method to counteract the object bias during training, which is detailed in the following section.

\vspace{-1em}
\section{Object Bias Alleviation}
\vspace{-0.5em}
\subsection{Problem Definition}
Given an image, an HOI detection model is expected to detect each interactive  triplet \Triplet{human}{verb}{object} and output their interaction score $s^{h,v,o}$, which is calculated as $s^{h,v,o} = s^v \cdot s^h \cdot s^o$.
$s^h$ and $s^o$ are the confidence scores for the detected person and object, respectively.
They are often obtained from the confidence score output by the detector. 
$s^v$ represents the verb score predicted by a classifier. 
In the following, we mainly consider the calculation of $s^v$ and omit the upper-script $v$ for notational convenience.

\vspace{-1em}
\subsection{Base Model}
\begin{figure*}[ht]
    \centering
    \includegraphics[width=1.0\textwidth]{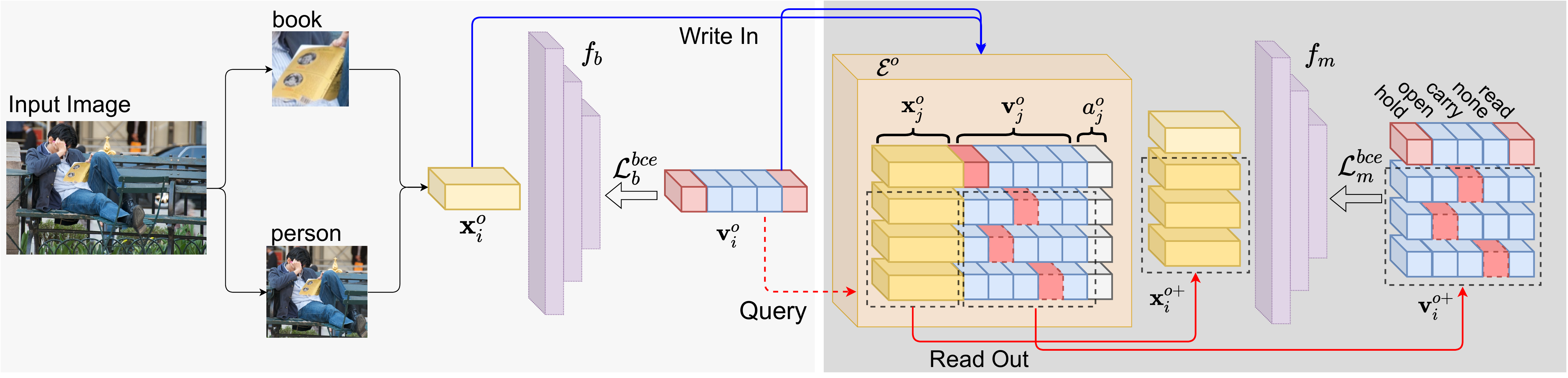}
    \vspace{-2em}
    \caption{\label{fig:method}
    Overview of the proposed method.
    Given an image, an HOI detection model extracts HOI features for each human-object pair. 
    A memory cell $\mathcal{E}^o$ is maintained for each object $o$.
    During training, instances are conditionally read and written into its respective cell with label-awareness.
    We show one human object pair ($o=\decorate{book}$) for clearance.
    }
    \vspace{-10pt}
\end{figure*}

In this work, we consider a generic HOI detection model, as shown in the left part of Fig.~\ref{fig:method}. 
It takes as input an image, detects all humans and objects, and links each human-object pair.
Thereafter, with message passing or context aggregation, a set of human-object pair representations, i.e., the HOI features $\{\mathbf{x}_i^{o}\}_{i=1}^N$ are obtained, where $N$ denotes the number of human-object pairs. 
Each feature $\mathbf{x}_i^{o}$ captures the interaction relation between a human and an object of class $o$.

We then feed these features into a classifier $f_b$ to predict verb scores: $\mathbf{s}_i = \sigma(f_b(\mathbf{x}_i^o))$, where $\sigma(\cdot)$ is a sigmoid function. 
Note that there can be multiple or no interactions within one human-object pair. 
Thus, the verb recognition is usually formulated as a multi-label classification problem.
The objective of the base model is formulated as follows:
\begin{equation}
\small
    \mathcal{L} = 
   \sum_{i=1}^N \mathcal{L}_{b}^{bce}(\mathbf{s}_i, \mathbf{v}_i^o) + \mathcal{L}_{aux},
\end{equation}
where $\mathbf{v}_i^o$ denotes the ground-truth label involving object $o$, $\mathcal{L}_{b}^{bce}$ is the binary cross entropy loss for verb classification and $\mathcal{L}_{aux}$ corresponds to other objectives of the base model such as interactiveness prediction and object localization.
As discussed before, the base model often severely suffers from the \textit{object bias problem}.
To overcome this issue, we design a novel Object-wise Debiasing Memory module which has minimal influence to the reasoning process of the base model and is plugable to any existing HOI detection methods.
\vspace{-1em}

\vspace{-2em}
\subsection{Object-wise Debiasing Memory}
It is widely accepted that instances from rare classes contain richer information for interaction understanding~\cite{NEURIPS2020_longtail_momentum,wu2020dbloss_ECCV20,Kang2020Decoupling}.
However, as discussed in Sec.~\ref{sec:bias_analysis}, frequent verb classes under each object dominate the prediction results, while other informative but rare ones are often ignored.
In view of this, 
we propose to re-sample HOI instances with a re-balancing strategy. 
In general, re-sampling has been shown to be an effective technique for class unbalance mitigation~\cite{wu2020dbloss_ECCV20,shen2016relaybp,li2019repair}.
However, in HOI detection, it is infeasible to directly apply these techniques.
On the one hand, the object bias problem is induced by object-conditional unbalance, rather than the overall one, which is distinct from the traditional class-imbalance scenario.
On the other hand, there can be multiple human-object interactions within a single image, simply re-sampling one image with rare classes may lead to oversampling of non-rare ones, which may further exacerbates the \textit{object bias problem}.

\begin{algorithm}[t]
\small

\caption{Read and Write Strategy for $\mathcal{E}^o$}
\label{algorithm:read_write}
\textit{\textbf{\textcolor[RGB]{128, 128, 128}{// Read Strategy}}}
\begin{algorithmic}
\Require
HOI instance $\{\mathbf{x}, \mathbf{v}\}$, number of required samples $k$

\Ensure
$k$ HOI instances and labels

\State features = [$\mathbf{x}$]; labels = [$\mathbf{v}$]
\While {number of sampled features \textless\ $k$} 
    \State \textcolor[RGB]{128, 128, 128}{// pick rare class entries when not selected}
    \State $j = argmax_{j}\ \sum_i dist($labels[i]$, \mathcal{E}^o[j])$ (Eq.~\ref{eq:hamming_dist})
    \State Append $\mathbf{x}_j, \mathbf{v}_j$ to features, labels
\EndWhile
\State return features, labels

\end{algorithmic}
\textit{\textbf{\textcolor[RGB]{128, 128, 128}{// Write Strategy}}}
\begin{algorithmic}
\Require
HOI instance \{$\mathbf{x}$, $\mathbf{v}$ \}, generation time $a$
\If {$\mathcal{E}^o$ is not full}
    \State Append $\{ \mathbf{x}, \mathbf{v}, a\}$ to $\mathcal{E}^o$
\Else:
    \If {$score(\mathbf{v}^{o}) \geq \tau^o$ (Eq.~\ref{eq:hamming_score})} 
        \State Replace entry of the longest duration with $\{ \mathbf{x}, \mathbf{v}, a\}$.
    \EndIf
\EndIf

\end{algorithmic}
\end{algorithm}

To circumvent this, we resort to the fine-grained feature-level re-sampling during model training.
Accordingly, we maintain a memory for each object, on which an effective read and write strategy is devised to operate.
We name this module Object-wise Debiasing Memory (ODM) and the framework is illustrated in Fig.~\ref{fig:method}.
Specifically, a memory cell $\mathcal{E}^o$ is maintained for each object $o$, which has a fixed size $n$ and stores three types of elements: the HOI feature $\mathbf{x}_j^{o}$, the verb label $\mathbf{v}_j^{o}$ and the feature generation time $a_j^{o}$.
During training, each ODM cell is sampled (read out) with label awareness, followed by a dynamic update (write in) operation to ensure feature consistency. 
The pseudo-code for read and write strategy is shown in Alg.~\ref{algorithm:read_write} and detailed as follows.

\vspace{-1em}
\subsubsection{Read Strategy}
To achieve verb balance under each object, it makes sense to assign high sampling priority to rare class instances.
At each training step, given an interactive HOI feature $\mathbf{x}_i^{o}$ with verb label $\mathbf{v}_i^{o}$, we take $\mathbf{v}_i^{o}$ as query and sample a set of $k$ HOI instances $\{\mathbf{x}_j^o, \mathbf{v}_j^o\}_{j=1}^k$ from the memory $\mathcal{E}^o$ such that the label distribution after sampling 
is less skewed.
To that end, we select from the memory with the largest weighted hamming distance, which  is calculated as:
\begin{equation}\label{eq:hamming_dist}
\small
    dist(\mathbf{v}_1^o, \mathbf{v}_2^o) = \sum_{t=1}^{c} w_t^o\cdot (\mathbf{v}_1^o[t] \oplus \mathbf{v}_2^o[t]),
\end{equation}
where $\oplus$ means XOR operation, $[\cdot]$ is subscription and $w_t^o$ is a weighting coefficient of the $t$-th class associated with object $o$.
Firstly, the hamming distance is employed to consider absent classes with respect to selected instance~\cite{gordo2013asymmetric}.
Secondly, the weighting mechanism ensures dynamic control over certain classes. 
Specifically, we calculate $w_t^o$ as $N_o/N_{v, o}$, where $N_o$ and $N_{v,o}$ denotes the number of object $o$ and interaction $\langle v, o \rangle$ in the training set.
By designing $w_t^o$ as inverse interaction frequency within object $o$,  rare class instances are prioritized and thus more frequently sampled from the memory.
In addition, we perform iterative sampling to avoid all selected samples are from the same class.
\vspace{-1em}
\subsubsection{Write Strategy}
During the writing stage, it is expected to store more rare class instances to ensure the sample complexity for memory reading.
Specifically, we treat one instance as write-feasible if its hamming score for a multi-hot label is greater than a threshold $\tau^o$.
The hamming score is given by:
\begin{equation} \label{eq:hamming_score}
\small
    score(\mathbf{v}_j^o) = \sum_{t=1}^c w_t^o \cdot \mathbf{v}_j^o[t],
\end{equation}
where $c$ is the number of verb classes and $w_t^o$ is the same weighting coefficient as that in Eq.~\ref{eq:hamming_dist}.
With this strategy, non-rare instances will not be written into the memory, thereby alleviating the risk of their dominance for model training.
When the memory is full, we replace the feature of the longest duration with write-feasible instances, so as to ensure timely update of memory contents.
\vspace{-1em}
\subsection{Training and Inference}

The proposed memory operations serve as an ad-hoc re-sampling approach to ensure more balanced training at each iteration.
After reading from the memory, we then leverage another classifier $f_m$ to perform more balanced interaction classification.
Inspired by recent work on class-imbalanced learning~\cite{zhou2020bbn,wang2020devil,Kang2020Decoupling}, we combine $f_m$ with the base classifier $f_b$ to achieve a trade-off between the debiasing and representation capability.
The overall objective is defined as follows:
\begin{equation}
\small
    \mathcal{L} = 
    \sum_{i=1}^{N}
    \mathcal{L}_{b}^{bce}(\mathbf{s}_i, \mathbf{v}_i^o) + \mathcal{L}_{m}^{bce}(\mathbf{s}_i^{+}, \mathbf{v}_i^{o+})
    + \mathcal{L}_{aux},
\end{equation}
where  $\mathbf{x}_i^{o+} = [\mathbf{x}_i^o; \{\mathbf{x}_j^o\}_{j=1}^k]$ and $\mathbf{v}_i^{o+} =[\mathbf{v}_i^o; \{\mathbf{v}_j^o\}_{j=1}^{k}]$ are obtained after the read operation and $\mathbf{s}_i^{+} = f_m\big(\mathbf{x}_i^{o+}\big)$.

During inference, given an HOI feature $\mathbf{x}_i$, we take the weighted combination of these two classifiers' output as the final prediction:
\begin{equation}
\small
    {\mathbf{\hat{v}}_i} = \sigma\big(\lambda f_b(\mathbf{x}_i) + (1 - \lambda) f_m(\mathbf{x}_i)\big),
\end{equation}
where $\lambda$ is a hyper-parameter balancing the two classifiers.
\vspace{-1.0em}
\section{Experiments} \label{sec:exps}
\vspace{-0.5em}
\subsection{Experimental Setting}
  
\noindent \textbf{Dataset} 
We conducted experiments on two benchmarks: HICO-DET~\cite{HORCNN_Chao2018WACV} and HOI-COCO~\cite{hou2021affordance_ATL_CVPR21}.
\textbf{HICO-DET} is the most widely employed benchmark in HOI detection.
It consists of 38,118 and 9,658 images in the training and test set, respectively.
HICO-DET covers the whole 80 object classes in MS-COCO~\cite{lin2014MSCOCO} and 117 verb classes, resulting in a total of 600 HOI categories in the form of \Triplet{person}{verb}{object}.
\textbf{HOI-COCO} is a recently introduced dataset based on V-COCO~\cite{gupta2015visual_VCOCO}.
It has a total of 9,915 images,
with 4,969 for training and 4,946 for test. 
There are 222 HOI categories composed of 21 verb classes from V-COCO and 80 MS-COCO object classes.


\noindent \textbf{Baselines}
As our goal is to prove the superiority and  versatility of the proposed method, we applied our approach to three existing methods: HOID~\cite{Wang_2020_CVPR_HOID}, SCG~\cite{zhang2021spatially_SCG_ICCV21} and QPIC~\cite{tamura2021qpic_CVPR21}.
\textbf{HOID} generates human-centric object proposal for interactive objects only. 
\textbf{SCG} is a recently proposed two-stage method leveraging spatial information for graph-based message propagation.
It achieves state-of-the-art performance with both fine-tuned and ground-truth detection among two-stage methods.
\textbf{QPIC} is an advanced one-stage method, which utilizes Transformer architecture to perform query-based detection and classification.

\noindent\textbf{Standard Evaluation Metrics}
We followed the standard evaluation setting \cite{HORCNN_Chao2018WACV} and reported mean average precision (mAP) for both datasets, where the mAP on rare (less than 10 training instances), non-rare and full classes are reported.
For both settings,
a prediction is regarded as positive if (1) the HOI classification is correct and (2) the detected human and object bounding boxes have IoUs greater than 0.5 with the ground-truth bounding box.

\noindent\textbf{Object-bias Evaluation Metric}
To quantitatively study how much object bias has been alleviated by our method,
we propose a new \textit{object bias} evaluation setting.
Specifically, we treated an interaction class as \textit{object rare} (\textit{object non-rare}) if $N_{v,o} / N_o <$ ($\geq$) $\alpha$.
On HICO-DET dataset, we set $\alpha$ to 0.3 based on its statistics.
Note that an originally non-rare class in the whole training set can be \textit{object rare} under this setting.
For each object, we computed the mean of Average Precision (AP) for \textit{object rare} and \textit{object non-rare} classes, respectively.
After that, we averaged across all objects to obtain mean Average Precision for \textbf{O}bject-\textbf{R}are (OR) and \textbf{O}bject-\textbf{N}on\textbf{R}are (ONR), respectively.
Besides, their \textbf{AVE}rage is also reported.
Different from the traditional evaluation, the \textit{object bias} evaluation protocol considers the performance within each object and thus offers a better test bed for quantifying a model's ability to overcome the \textit{object bias} problem. 

\vspace{-1em}
\subsection{Object Bias Evaluation} \label{exp:obj_bias}
\begin{table}[t]
    \centering
    \small
    \caption{\label{table:hicodet_objrare}
        Performance comparison under object-bias setting on HICO-DET.
        OR and ONR denote \textbf{O}bject-\textbf{R}are and \textbf{O}bject-\textbf{N}on\textbf{R}are, respectively.
    }
    \scalebox{0.9}{
    \begin{tabular}{c||cc|cc||cc|cc||cc}
    \toprule
    \textbf{Detector} & \multicolumn{4}{c||}{\textit{Pre-trained Detector}} & \multicolumn{4}{c||}{\textit{Fine-tuned Detector}} & \multicolumn{2}{c}{\textit{Oracle Detector}}    \\
    \midrule
    \midrule
    \textbf{Method} &  HOID~\cite{Wang_2020_CVPR_HOID}    &   +Ours   &   SCG~\cite{zhang2021spatially_SCG_ICCV21}     &   +Ours   &   QPIC~\cite{tamura2021qpic_CVPR21}    &   +Ours   &   SCG    &   +Ours   &   SCG     &   +Ours \\
    \midrule
    \textbf{OR}  &  17.05   &   \textbf{19.02}   &   18.38   &   \textbf{19.47}   &   26.29   &   \textbf{26.96}   &   28.67   &   \textbf{30.21}   &   51.03   &   \textbf{52.48} \\
    \textbf{ONR} &  24.24   &   \textbf{24.33}   &   \textbf{25.06}   &   25.01   &   34.64   &   \textbf{34.65}   &   40.72   &   \textbf{41.08}   &   73.97   &   \textbf{75.43} \\
    \textbf{AVE} &  20.65   &   \textbf{21.17}   &   21.72   &   \textbf{22.24}   &   30.47   &   \textbf{30.81}   &   34.69   &   \textbf{35.64}   &   62.50   &   \textbf{63.95} \\
    \bottomrule
    \end{tabular}}
    \vspace{-20pt}
\end{table}

\noindent The results under the new \textit{object bias} setting are shown in Tab.~\ref{table:hicodet_objrare}.
With SCG as baseline, our method significantly improves \textit{object rare} classes by a clear margin of +\textbf{1.09} mAP, +\textbf{1.54} mAP and +\textbf{1.45} mAP under three detection settings.
Besides, the proposed method also boosts HOID and QPIC by +\textbf{0.97} mAP and +\textbf{0.67} mAP under OR setting.
This provides evidence that our method can effectively alleviate the object bias problem.
Notably, the proposed module can also improve ONR classes in most cases.
\vspace{-1em}
\begin{table}
\begin{minipage}[t]{.48\textwidth}
    \centering
    \caption{\label{hicodet_pretrained}
        Results on HICO-DET with \textit{pre-trained} detector.
    }
    \begin{tabular}{l ccc}
    \toprule
    \textbf{Method} & \textbf{Full} & \textbf{Rare} & \textbf{Non-rare} \\
    \midrule
    iCAN~\cite{gao2018ican_BMVC18}                          & 14.84 & 10.45 & 16.15  \\
    TIN~\cite{li2019transferable_TIN_CVPR19}                & 17.03 & 13.42 & 18.11  \\
    DRG~\cite{gao2020drg_ECCV20}                            & 19.26 & 17.74 & 19.71  \\
    VCL~\cite{hou2020visual_VCL_ECCV20}                     & 19.43 & 16.55 & 20.29  \\
    ACP~\cite{kim2020detecting_ACP_ECCV}                    & 20.59 & 15.92 & 21.98  \\
    DJ-RN~\cite{li2020djrn_CVPR20}                          & 21.34 & \textbf{18.53} & 22.18  \\
    \midrule
    HOID*~\cite{Wang_2020_CVPR_HOID}                        &    19.58 & 15.29 & 20.96 \\
    +Ours                                                   &    20.45 & 16.18 & 21.73 \\
    \midrule
    SCG*~\cite{zhang2021spatially_SCG_ICCV21}               & 20.99 & 16.30 & 22.40 \\
    +Ours                                                   & \textbf{21.50} & 17.59 & \textbf{22.67} \\
    \bottomrule
    \end{tabular}
\end{minipage}\qquad
\hspace{\fill}
\begin{minipage}[t]{.48\textwidth}
    \centering
    \caption{\label{hicodet_finetuned}
        Results on HICO-DET with \textit{fine-tuned} detector.
    }
    \begin{tabular}{l ccc}
    \toprule
    \textbf{Method} & \textbf{Full} & \textbf{Rare} & \textbf{Non-rare} \\
    \midrule
    PPDM~\cite{liao2020ppdm_CVPR20}                         & 21.73 & 13.78 & 24.10 \\
    HOI-Trans~\cite{zou2021end_HOITransformer_CVPR21}       & 23.46 & 16.91 & 25.41 \\
    ATL~\cite{hou2021affordance_ATL_CVPR21}                 & 27.68 & 20.31 & 29.89 \\
    AS-Net~\cite{chen2021reformulating_ASNet_CVPR21}        & 28.87 & 24.25 & 30.25 \\
    FCL~\cite{hou2021detecting_FCL_CVPR21}                  & 29.12 & 23.67 & 30.75 \\
    \midrule
    QPIC* ~\cite{tamura2021qpic_CVPR21}                     & 29.04 & 21.55 & 31.27 \\
    QPIC + Ours                                             & 29.26 & 22.07 & 31.41 \\
    \midrule
    SCG* ~\cite{zhang2021spatially_SCG_ICCV21}                & 31.08 & 24.14 & 33.15 \\
    SCG + Ours                                                & \textbf{31.65} & \textbf{24.95} & \textbf{33.65}   \\
    \bottomrule
    \end{tabular}
\end{minipage}
\vspace{-2.4em}
\end{table}

\subsection{Standard Evaluation}
\noindent\textbf{Results on HICO-DET}
We followed ~\cite{zhang2021spatially_SCG_ICCV21} to report the results with \textit{detector pre-trained on MS-COCO~\cite{lin2014MSCOCO}} (HOID and SCG), \textit{detector find-tuned on HICO-DET} (SCG and QPIC) and \textit{oracle detector} (SCG).  
The results can be found in Tab.~\ref{hicodet_pretrained},~\ref{hicodet_finetuned} and~\ref{hicodet_oracle}, respectively.

Our method improves the performance of all three baseline methods across all detection settings.
For instance, with pre-trained detector, our method promotes HOID and SCG by +\textbf{0.89} and +\textbf{1.29} mAP on rare classes, respectively, which amounts to \textbf{6}\% and \textbf{8}\% relative improvements.
When leveraging fine-tuned detector, the proposed approach can improve QPIC and SCG on rare classes by +\textbf{0.52} and +\textbf{0.81} mAP.
In particular, with the detection quality improved, our method also enhances non-rare classes by a noticeable margin.
Lastly, with the oracle detector, the proposed method can advance SCG on both rare (+\textbf{1.22} mAP) and non-rare classes (+\textbf{1.26} mAP).
These results demonstrate the superiority of the proposed method.
As a side product, we achieve new state-of-the-art on the HICO-DET dataset.
\vspace{-1.0em}

\begin{table}[htb]
\vspace{-2em}
\begin{minipage}[t]{.48\textwidth}
    \centering
    \caption{\label{hicodet_oracle}
        Results on HICO-DET with \textit{oracle detector}.
    }
    \begin{tabular}{l ccc}
    \toprule
    \textbf{Method} & \textbf{Full} & \textbf{Rare} & \textbf{Non-rare} \\
    \midrule
    iCAN~\cite{gao2018ican_BMVC18}                          & 33.38 & 21.43 & 36.95 \\
    TIN~\cite{li2019transferable_TIN_CVPR19}                & 34.26 & 22.90 & 37.65 \\
    Peyre \textit{et al.}~\cite{peyre2019analogies_ICCV19} & 34.35 & 27.57 & 36.38 \\
    FCL~\cite{hou2021detecting_FCL_CVPR21}                  & 44.26 & 35.46 & 46.88 \\
    \midrule
    SCG*~\cite{zhang2021spatially_SCG_ICCV21}               & 51.03 & 38.93 & 54.65 \\
    SCG + Ours                                              & \textbf{52.29} & \textbf{40.15} & \textbf{55.91} \\
    \bottomrule
    \end{tabular}
\end{minipage}\qquad
\begin{minipage}[t]{.48\textwidth}
    \centering
    \caption{\label{hoicoco_main}
        Results on the HOI-COCO.
        * indicates reproduced baseline.
    }
    \begin{tabular}{l ccc}
    \toprule
    \textbf{Method} & \textbf{Full} & \textbf{Rare} & \textbf{Non-rare} \\
    \midrule
    Baseline~\cite{hou2021affordance_ATL_CVPR21} & 22.86 &  6.87    & 35.27 \\
    +VCL~\cite{hou2020visual_VCL_ECCV20}     &   23.53   &   8.29    & 35.36  \\
    +ATL~\cite{hou2021affordance_ATL_CVPR21} &   23.40   &   8.01    &   35.34   \\
    \midrule
    Baseline*                                &   22.87   &   6.98    &   35.21   \\
    +CDN~\cite{zhang2021mining_CDN_NIPS21}                                     &   23.15   &   7.25    &   \textbf{35.49}   \\
    +Ours                                    &   \textbf{23.73}   &   \textbf{8.58}    &   \textbf{35.49}   \\
    \bottomrule
    \end{tabular}
\end{minipage}
\vspace{-2.5em}
\end{table}

\noindent\textbf{Results on HOI-COCO}
We followed ~\cite{hou2021affordance_ATL_CVPR21} to provide results with MS-COCO pre-trained detector, which is the most typical setting for two-stage methods.
The results are shown in Tab.~\ref{hoicoco_main}.
For fair comparison, we reproduced the baseline method used in ATL~\cite{hou2021affordance_ATL_CVPR21}. 
We also compared with the debiasing technique applied in CDN~\cite{zhang2021mining_CDN_NIPS21}, which aims to alleviate the general long-tail problem.
It can be observed that our method outperforms these debiasing methods on this relatively small scale dataset, especially for rare classes.


\vspace{-1em}
\subsection{Ablation Studies} 
We studied the effectiveness of our proposed method.
All experiments are conducted on HICO-DET dataset with the SCG~\cite{zhang2021spatially_SCG_ICCV21} baseline, and evaluated under both standard protocol and the proposed \textit{object bias} setting.

\begin{table}[h]
    \centering
    \caption{Comparsion with debiasing methods.}\label{tab:debias}
    \scalebox{1.0}{
    \begin{tabular}{c|l||c c c|c c c}
         \toprule
        Type & Method & \textbf{Full}  & \textbf{Rare} & \textbf{Non-rare}  & \textbf{OR} & \textbf{ONR} & \textbf{AVE} \\
        \midrule
         & Baseline    &        20.99 & 16.30 & 22.40  & 18.38 & \textbf{25.06} & 21.70 \\
        \midrule
         \multirow{4}{*}{Reweighting} & +inv. freq.        & 17.58 & 14.15 & 18.61 &   9.77 & 21.01 & 15.39 \\
         & +CB-Loss($0.9999$)~\cite{cui2019class}        
                            & 14.30 & 13.54 & 14.53 &   9.93   &   21.48   &   15.71 \\
         & +CB-Loss($0.999$)~\cite{cui2019class}         
                            & 13.34 & 12.96 & 13.45 &   9.02      & 20.73   &   14.88 \\
         & +CB-Loss($0.99$)~\cite{cui2019class}  
                            & 13.98 & 13.20 & 14.21 &   9.46   &   20.98   &   15.22 \\
        \midrule
         General & +AT~\cite{wang2020benchmarkbias}&              20.49   &   16.22   &   21.77   &   18.12   &   24.46   &   21.29   \\
         Debiasing & +DIT~\cite{wang2020benchmarkbias}           &  18.13    &   16.99   &   18.47  &   17.35   &   23.05   &  20.20   \\
         \midrule
          SGG    & +{TDE}~\cite{Tang_2020_CVPR_unbiased_SGG}&
                              20.44    &   14.89   &   22.10   &   18.30   &   24.44   &   21.37 \\
         Debiasing & +{PCPL}~\cite{yan2020pcpl}              
                            &   16.93   &   12.95   &   18.12   &   15.04   &   24.27   &   19.65   \\
         \midrule

         & +Ours              & \textbf{21.50} & \textbf{17.59} & \textbf{22.67} & \textbf{19.47} & 25.01 & \textbf{22.24} \\
         \bottomrule
    \end{tabular}
    }
    \vspace{-1.5em}
\end{table}
\vspace{-1.5em}
\subsubsection{Comparison with other Debiasing Methods}
We compared our method with various debiasing methods in Tab.~\ref{tab:debias}. 
The competitors include loss re-weighting methods, general debiasing methods and Scene Graph Generation (SGG) debiasing methods. 
We observed all these methods degrade the original baselines.
This may be related to the strong interference with the original training process.
Besides, some methods are designed to tackle the globally long-tail problem and single-label classification, thus incapable of resolving the object-conditional long-tail problem in HOI detection.

\vspace{-1.0em}
\subsubsection{Efficacy of Classifiers}
\begin{table}[!t]
    \centering
    \small
    \caption{\label{table:ablation_classifier}
        Performance of different classifiers on HICO-DET.
    }
	\adjustbox{}{
    \begin{tabular}{c|c|ccc|ccc}
    \toprule
    \textbf{Detector} & \textbf{Classifier}             &   \textbf{Full}    &   \textbf{Rare}    & \textbf{Non-rare}      
                                &   \textbf{OR}      &   \textbf{ONR}     & \textbf{AVE} \\
    \midrule
    \multirow{3}{*}{Pre-trained on MS-COCO} & $f_b$                           
                                    &   21.08   &   16.66    &   22.40      
                                    &   19.00   &   24.38    &   21.69              \\
    & $f_m$                           &   20.79   &   16.50    &   22.08      
                                    &   18.59   &   24.87    &   21.73              \\
    & $full$                          & \textbf{21.50}   &   \textbf{17.59}    &   \textbf{22.67}   
                                    & \textbf{19.47}   &   \textbf{25.01}    &   \textbf{22.24} \\
    \midrule
    \multirow{3}{*}{Fine-tuned on HICO-DET}& $f_b$                       &   31.24   &   24.77   &   33.17   
                                &   30.04   &   40.62   &   35.33   \\
    & $f_m$                       &   30.60   &   23.30   &   32.77   
                                &   29.43   &   40.79   &   35.11   \\
    & $full$                        &   \textbf{31.65}   &   \textbf{24.95}    &   \textbf{33.65}   
                                &   \textbf{30.21}   &   \textbf{41.08}    &   \textbf{35.64} \\
    \midrule
    \multirow{3}{*}{Oracle} & $f_b$                       &   51.24   &   39.27   &   54.82   
                                &   51.81   &   74.73   &   63.27   \\
    & $f_m$                       &   51.09   &   37.94   &   55.01   
                                &   51.28   &   75.15   &   63.21   \\
    & $full$                        &   \textbf{52.29}   &   \textbf{40.15}    &   \textbf{55.91}   
                                &   \textbf{52.48}   &   \textbf{75.43}    &   \textbf{63.95}\\
    \bottomrule
    \end{tabular}}
    \vspace{-2em}
\end{table}
\vspace{-0.6em}
The distinctive importance of the verb classifier in the base model ($f_b$), the one trained with ODM ($f_m$) and the full classifier $(\lambda f_b + (1-\lambda)f_m)$ are explored in this experiment. From the results in Tab.~\ref{table:ablation_classifier}, we see that
for all three detectors, $f_m$ is inferior to $f_b$ on both evaluation protocols.
However, when combining these two together, the final performance can be further promoted.
This is mainly because these two classifiers focus on different classes and are in fact complementary to each other.

\vspace{-1.0em}
\begin{figure}
\begin{minipage}[t]{0.48\textwidth}
    \flushleft
    \centering
    \includegraphics[width=1.0\textwidth, trim=0 0 0 0, clip]{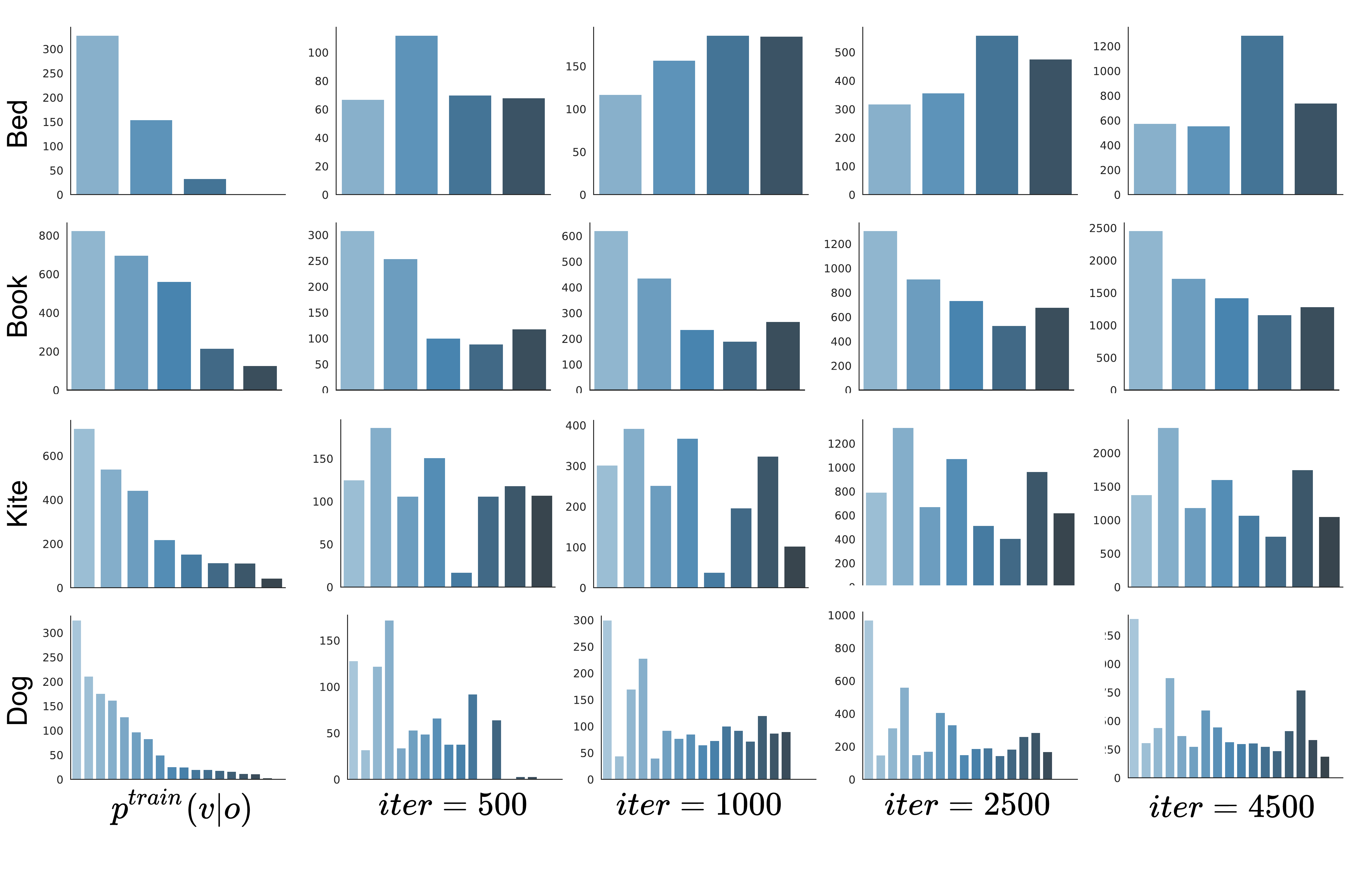}
    \vspace{-10pt}
    \caption{\label{fig:memvis}
    The evolution of accumulated verb distribution after reading from the proposed ODM for 4 randomly selected objects. 
    The leftmost column shows $p^{train}(v|o)$ and the other 4 columns represent the sampled verb distribution at different iterations.
    }
    \vspace{-15pt}
\end{minipage}
\hfill
\begin{minipage}[t]{0.48\textwidth}
    \flushright
    \centering
    \includegraphics[width=1.0\textwidth, trim=0 0 0 0, clip]{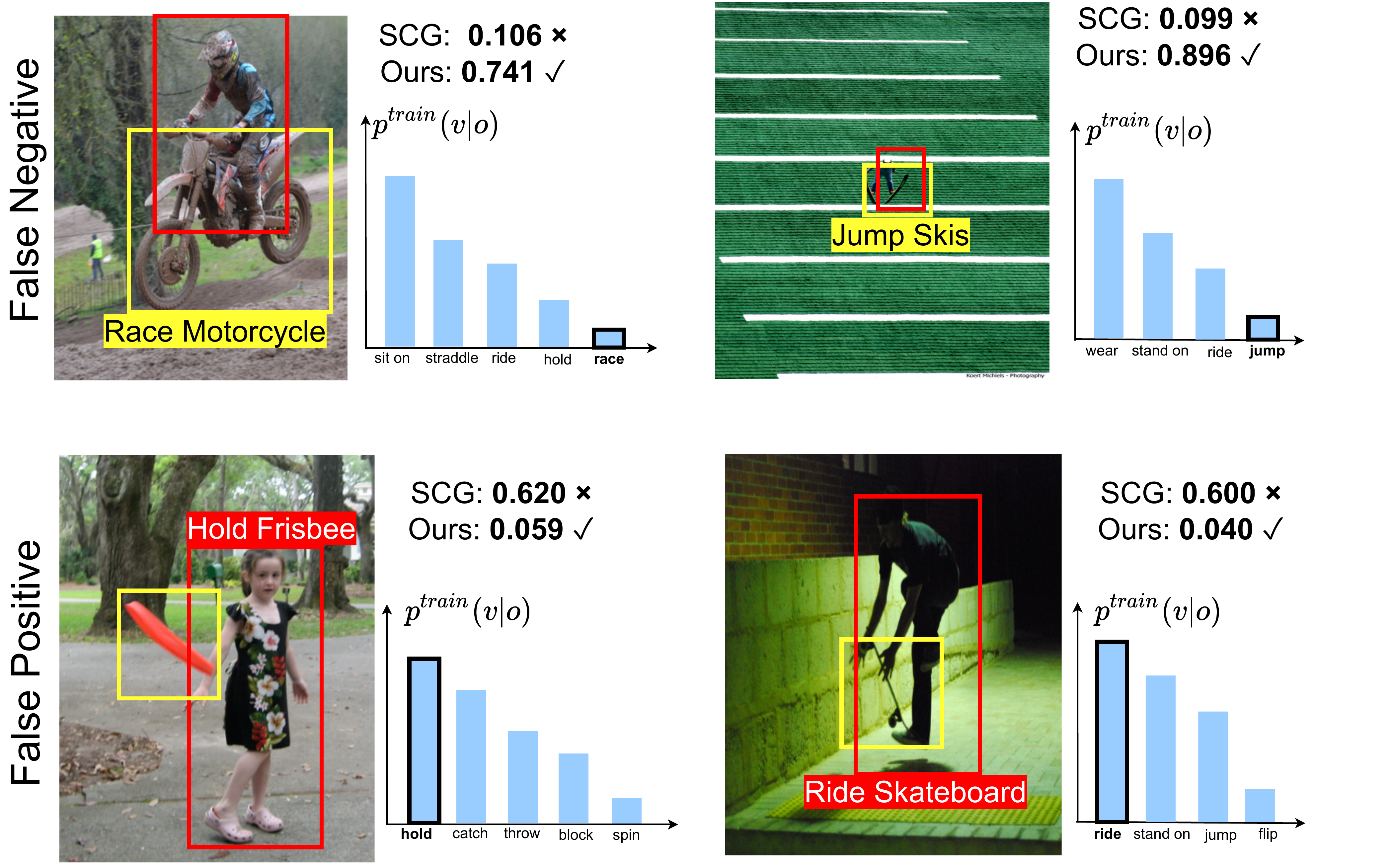}
    \caption{\label{fig:quali}
    False negative (top two) and false positive (bottom two) instances from the SCG baseline on HICO-DET test set.
    For each instance, the $p^{train}(v|o)$ is also shown, where the involved verb is bold by a rectangular.
    }
    \vspace{-2.0em}
\end{minipage}
\end{figure}
\subsection{Visualizations}
\vspace{-0.5em}
\subsubsection{Effects of Memory}
We studied how the proposed ODM alleviates the distribution imbalance and illustrated the evolution of verb distribution after reading from the memory in Fig.~\ref{fig:memvis}.
With these examples, we can conclude that our method can effectively address the label imbalance problem under each object.
Especially, at the 2500-$th$ iteration, the verb distribution is already less skewed, which remains stable till the end of this epoch ($\sim$4.5k iterations).
\vspace{-1.2em}
\subsubsection{Qualitative Results}
We show some qualitative results in Fig.~\ref{fig:quali}. 
For the two false negative instances (top two), the baseline model assigns low score to ground-truth interactions, wherein both involved verbs are conditionally rare in the training set  (\decorate{race} for \decorate{motorcycle}, \decorate{jump} for \decorate{skis}).
For the false positive instances (bottom two), the baseline favors more frequent verbs (\decorate{hold} for \decorate{frisbee}, \decorate{ride} for \decorate{skateboard}), though the interaction prediction is incorrect.
In contrast, our method can overcome these two kinds of errors and achieve better performance.


\vspace{-1.5em}
\section{Conclusion and Future Work}
\label{sec:conclusion}
\vspace{-0.8em}
In this work, we systematically studied the \textit{object bias} problem in Human-Object Interaction detection.
We demonstrated the recognition of this problem from the aspects of unbalanced label distribution and biased model learning, and advocated a new protocol to comprehensively evaluate model performance.
To reduce the heavily skewed label distribution under each object, we proposed an Object-wise Debiasing Memory to facilitate balanced sampling of HOI instances.
Extensive experiments validate the effectiveness of the proposed method,
demonstrating that it can significantly alleviate the \textit{object bias problem} and outperform advanced baselines with large margins.
Due to the universal existence of the \textit{bias problem}, in the future, we plan to explore identifying bias factors in other related tasks such as visual relation detection and scene graph generation. 

\vspace{-1.2em}
\section*{Acknowledgement}
\vspace{-1.0em}
This research is supported by the National Research Foundation, Singapore under its Strategic Capability Research Centres Funding Initiative. 
Any opinions, findings and conclusions or recommendations expressed in this material are those of the author(s) and do not reflect the views of National Research Foundation, Singapore.
 
\clearpage

\bibliographystyle{splncs04}
\bibliography{egbib}
\end{document}


\pagestyle{headings}
\mainmatter
\def\ECCVSubNumber{4206}  

\title{Chairs Can be Stood on: Overcoming Object Bias in
Human-Object Interaction Detection \\ Supplementary Material} 

\titlerunning{ECCV-22 submission ID \ECCVSubNumber} 
\authorrunning{ECCV-22 submission ID \ECCVSubNumber} 
\author{Anonymous ECCV submission}
\institute{Paper ID \ECCVSubNumber}

\maketitle



\section{Additional Experiments}
\subsection{Known Object Setting}
Following previous work~\cite{HORCNN_Chao2018WACV,li2020hoi_IDN,zhang2021spatially_SCG_ICCV21}, we also report results on the HICO-DET dataset~\cite{HORCNN_Chao2018WACV} with Known Object (KO) setting in Tab.~\ref{table:hicodet_ko_pre} and Tab.~\ref{table:hicodet_ko}, respectively.
It can be observed that our method surpasses the baselines under this setting.
\begin{table}[]
    \vspace{-3em}
    \begin{minipage}{0.49\textwidth}
    \centering
    \caption{\label{table:hicodet_ko_pre}
        Performance comparison on HICO-DET under the Known Object (KO) setting with \textit{pre-trained} detector.
    }
    \adjustbox{}{
    \begin{tabular}{lccc}
    \toprule
    \textbf{Method} & \textbf{Full} & \textbf{Rare} & \textbf{Non-rare} \\
    \midrule
    \midrule      
    iCAN~\cite{gao2018ican_BMVC18}                          & 16.26 & 11.33 & 17.73  \\
    TIN~\cite{li2019transferable_TIN_CVPR19}                & 19.17 & 15.51 & 20.26  \\
    DRG~\cite{gao2020drg_ECCV20}                            & 23.40 & 21.75 & 23.89  \\
    VCL~\cite{hou2020visual_VCL_ECCV20}                     & 22.00 & 19.09 & 22.87  \\
    DJ-RN~\cite{li2020djrn_CVPR20}                          & 23.69 & 20.64 & 24.60  \\
    \midrule
    SCG*~\cite{zhang2021spatially_SCG_ICCV21}               & 24.53 & 20.00 & 25.88 \\
    SCG + Ours                                              & \textbf{25.54} & \textbf{21.93} & \textbf{26.61} \\
    \bottomrule
    \end{tabular}}\hfill
    \end{minipage}
    \begin{minipage}{0.48\textwidth}
    \centering
    \caption{\label{table:hicodet_ko}
        Performance comparison on HICO-DET under the Known Object (KO) setting with \textit{fine-tuned} detector.
    }
    \adjustbox{}{
    \begin{tabular}{lccc}
    \toprule
    \textbf{Method} & \textbf{Full} & \textbf{Rare} & \textbf{Non-rare} \\
    \midrule
    \midrule      
    PPDM~\cite{liao2020ppdm_CVPR20}                         & 24.58 & 16.65 & 26.84 \\
    HOI-Trans~\cite{zou2021end_HOITransformer_CVPR21}       & 26.15 & 19.24 & 28.22 \\
    ATL~\cite{hou2021affordance_ATL_CVPR21}                 & 27.38 & 22.09 & 28.96 \\
    AS-Net~\cite{chen2021reformulating_ASNet_CVPR21}        & 31.74 & 27.07 & 33.14 \\
    FCL~\cite{hou2021detecting_FCL_CVPR21}                  & 31.31 & 25.62 & 33.02 \\
    \midrule
    SCG* ~\cite{zhang2021spatially_SCG_ICCV21}                & 33.74   & 26.41   & 35.95 \\
    SCG + Ours                                                & \textbf{34.52}   & \textbf{27.34}   & \textbf{36.67}   \\
    \bottomrule
    \end{tabular}
    }
    \end{minipage}
    \vspace{-3em}
\end{table}

    

\begin{table}[t]
    \centering
    \small
    \caption{\label{table:lambda}
        Performance comparison with different $\lambda$. 
    }
    \adjustbox{}{
    \begin{tabular}{c|ccc|ccc}
    \toprule
    $\lambda$       &   \textbf{Full}    &   \textbf{Rare}    & \textbf{Non-rare} 
                    &   \textbf{OR}      &   \textbf{ONR}     & \textbf{AVE}            \\
    \midrule
    0.2              &   21.32    &   17.39    &   22.50       
                     &   19.33    &   24.66    &   22.00                                \\ 
    0.4              &   \textbf{21.50}    &   \textbf{17.59}    &   \textbf{22.67}       
                     &   \textbf{19.47}    &   \textbf{25.01}    &   \textbf{22.14}     \\   
    0.6              &   21.41    &   17.44    &   22.60       
                     &   19.33    &   24.98    &   22.16                                \\
    0.8              &   21.17    &   17.10    &   22.38       
                     &   19.02    &   25.01    &   22.02                                \\
    \bottomrule
    \end{tabular}}
    \vspace{-10pt}
\end{table}
\vspace{-1em}
\subsection{Hyper-parameter Analysis}
The detailed analysis of the coordination of these two classifiers with respect to $\lambda$ is shown in Tab.~\ref{table:lambda}.
It can be seen both classifiers are essential for the performance improvement.

\vspace{-1em}
\subsection{Efficiency and Memory Comparison}
\begin{table}[h]
    \small
    \centering
    \caption{Efficiency and memory comparison.}
    \begin{tabular}{l|c|c|c|c}
        \toprule
          & \textbf{train/img} & \textbf{test/img} & \textbf{\#param (train)} & \textbf{\#param (test)} \\
        \midrule
        SCG~\cite{zhang2021spatially_SCG_ICCV21}  & 428.16ms & 248.50ms & 16.04M & 16.04M \\ 
        +Ours & 440.72ms & 251.32ms      & 17.12M & 16.50M \\
        \bottomrule
    \end{tabular}
    \label{tab:time_size}
\end{table}

We compare the memory and computational cost with SCG~\cite{zhang2021spatially_SCG_ICCV21} in Tab.~\ref{tab:time_size}.
Note that the adopted detector (\textit{i.e.}, Faster R-CNN~\cite{ren2015fasterrcnn}) is not counted as the detection results can be obtained via one-pass inference for all images  before training.
It can be observed the overhead brought by our method is negligible for both training and test.

\vspace{-2.0em}
\section{Implementations}
\vspace{-0.5em}
\subsection{Overall Implementations}
We conducted all experiments on 4 Nvidia 2080Ti GPUs. 
Due to resource limitation, we reduced the batch size of SCG and QPIC to 8 and linearly scaled their learning rate.\footnote{This may slightly influence the performance and result in inconsistency between reported and reproduced ones.}
For QPIC, we started from a trained model and finetuned it with the proposed method for a total of 15 epochs. 
The learning rate is decayed by 0.1 at the 10-th epoch. 
For the other two baselines, we followed the default scheduling and started the training of $f_m$ from the $3rd$ epoch for stability.
$\lambda$ is empirically set to 0.4 in all experiments.

For the proposed method, we parameterize $f_m$ as a three layer Multi-layer Perceptron with ReLU activation function. 
For each memory cell, we set the size $n$ to 16 for each object and $k$ to 4.
About the write operation, $\tau^o$ is set to the third smallest $w^o$ (for objects with more than 5 associated verbs) or 0 (other objects).
Regarding to other aspects with respect to base models (feature extractor, sampling strategy, and loss function), we adopted their default settings.
More details are as follows.

\begin{table*}
    \centering
    \caption{\label{table:imp_detail}
        Implementation details of baseline methods
    }
    \adjustbox{}{
    \begin{tabular}{l|ccc}
    \toprule
    Method      &   SCG~\cite{zhang2021spatially_SCG_ICCV21}    &   HOID~\cite{Wang_2020_CVPR_HOID}    & QPIC~\cite{tamura2021qpic_CVPR21}                           \\
    \midrule
    Venue       &   ICCV'21       &   CVPR'20       &   CVPR'21         \\
    Batch Size (default)   &   $4\times8$    &   $4\times4$    &   $2\times8$      \\
    Batch Size (ours)  &   $2\times4$    &   $4\times4$    &   $2\times4$      \\
    Feature Extractor
                &   ResNet50-FPN   &  ResNet50-FPN   &   ResNet-50  \\
    Proposal    Generation
                &   Faster-RCNN~\cite{ren2015fasterrcnn}    &   HO-RPN~\cite{Wang_2020_CVPR_HOID}    &   QPIC~\cite{tamura2021qpic_CVPR21}/DETR~\cite{carion2020detr}        \\
    Interaction Loss 
                &   BCE+focal~\cite{lin2017focal}    &   BCE    &   BCE        \\
    Interactiveness Score~\cite{li2019transferable_TIN_CVPR19}
                &   Yes           &   No            &   No       \\
    Low-grade Suppression~\cite{li2019transferable_TIN_CVPR19}
                &   Yes           &   No            &   No       \\
    Sample Strategy 
                &   $\#$human\&$\#$object    &   pos/neg ratio & None  \\
    
    \bottomrule
    \end{tabular}}
\end{table*}

\subsection{Implementation of Baselines} 
The implementation details of the baselines are listed in Tab.~\ref{table:imp_detail}. 
In this table, the batch size is represented as \textit{number of images per GPU} $\times$ \textit{number of GPUs}.
BCE stands for binary cross-entropy loss.
Kindly find the codes with the corresponding model names in the zip file.

\subsection{More Hyper-parameter Settings}
\begin{table}[h]
    \small
    \centering
    \caption{Model performance of different hyper-parameter settings.}\label{tab:lrbs}
    \begin{tabular}{c|l|c|c|c}
         \toprule
         \textbf{Index} & \textbf{Setting}   & \textbf{Full}\uparrow & \textbf{Rare}\uparrow & \textbf{Non-rare}\uparrow  \\
         \midrule
         1  & Baseline (4 GPUs * 2 image, unscaled lr) & 19.94 & 14.70 & 21.50 \\ 
         \midrule
         2  & Baseline (4 GPUs * 1 image, scaled lr) & 20.75 & 15.96 & 22.18 \\
         3  & + Ours                   & \textbf{21.16} & \textbf{17.41} & \textbf{22.28} \\
        \midrule
         4  & Baseline (4 GPUs * 2 image, scaled lr)   & 20.99 & 16.30 & 22.40 \\
         5  & + Ours                    & \textbf{21.50} & \textbf{17.59} & \textbf{22.67} \\
        \bottomrule
    \end{tabular}
\end{table}
Due to resource limitation, we used a smaller batch size and scaled learning rate for both the baseline (SCG) and our method in all previous experiments.
We also studied the performance of baseline and our method under other hyper-parameter settings in Tab.~\ref{tab:lrbs}.
It can be observed that a) smaller batch size results in worse performance (2\&4, 3\&5). b) linearly scaling learning rate with respect to batch size can prohibit performance degradation to some degree (1\&4).
c) Under different training settings, our method outperforms the baseline by a considerable margin (2\&3, 4\&5).

\vspace{-1em}
\section{Discussion on Debiasing Baselines}
\noindent\textbf{Re-weighting Methods}
For re-weighting methods (\textit{i.e.}, inverse frequency weighting and CB-Loss~\cite{cui2019class}), we followed their conventions and computed the number of HOI instances (\textit{i.e.} interactive human-object pairs) in the training set to facilitate the weight calculation.
However, this leads to severe performance degradation.
We conjecture that there are mainly two reasons.
Firstly, these loss functions are all designed for reducing the general bias, instead of the \textit{object bias} studied in this paper. 
Secondly, these re-weighting strategies interfere a lot the original training process, which requires complex interaction recognition and reasoning.
In contrast, our proposed method allows dynamic adjustment with respect to each HOI instance in the training process, thereby improving the performance. 

\noindent\textbf{General Debiasing Methods}
For Adversarial Training (AT)~\cite{wang2020benchmarkbias}, we trained the model with another classifier, whose output dimension equals to the number of object classes (\textit{i.e.}, 80 in HICO-DET). 
Given each human-object feature, a cross-entropy loss with a flat label, \textit{i.e.}, $\mathbf{1}/N_o$ is introduced to the original training process, so that the representation is expected to be object-agnostic.
For Domain Independent Training (DIT)~\cite{wang2020benchmarkbias}, we trained the model with another classifier. The output dimension of this classifier equals to the number of total interactions (\textit{i.e.}, 600 in HICO-DET).
During inference, the interaction prediction is taken as the maximum probability over all interactions involving this verb.
We observe significant performance degradation with these methods. 
The key reason to this is that these methods ignore the object factor in their representations, which is essential for interaction recognition.

\noindent\textbf{SGG Debiasing Methods}
The original TDE~\cite{Tang_2020_CVPR_unbiased_SGG} aims to alleviate the contextual bias in Scene Graph Generation (SGG).
Besides the original forward pass, it conducts a second forward pass in the same model by masking (\textit{e.g.}, set to zero) both the subjects and the objects.
The final prediction is taken as the subtraction between the original logits and the logits in the second pass.
In this way, the biasing effects caused by factors other than the subject and object are expected to be eliminated.
In this work, to alleviate the object bias, we conduct a second forward pass by masking everything other than the object.
Then, similarly, the final output logits is obtained by subtracting this logits from the original ones.
By doing the subtraction, the output is expected to be less affected by the object bias problem, following the intuition of ~\cite{Tang_2020_CVPR_unbiased_SGG}.
In PCPL~\cite{yan2020pcpl}, we take the representation of an HOI class as the average of all features that involve this interaction class.
We argue that the failure of these methods may result from the ignorance of multi-label setting, which results in different logits in TDE (since single-label classification is conducted for SGG.) and imprecise class embedding estimation in PCPL (because an embedding for one instance may be counted into multiple classes, confusing the representations).



\section{More Visualizations}
\subsection{More Memory Evolutions}
\begin{figure}[t]
    \centering
    \includegraphics[width=1.0\columnwidth, trim=0 0 0 0, clip]{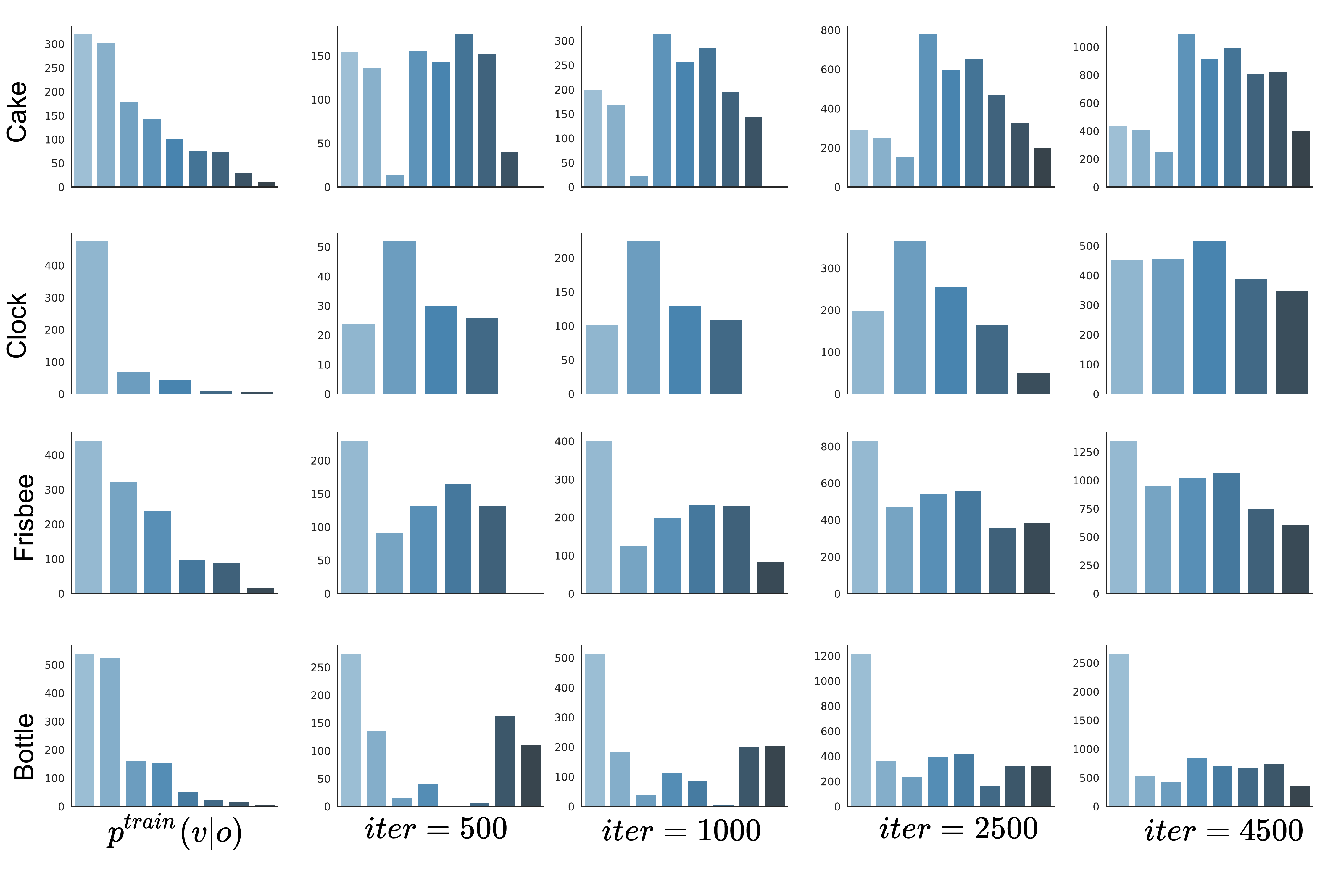}
    \vspace{-2em}
    \caption{\label{fig:memvis}
    Extra examples for the evolution of accumulated verb distribution after reading from the proposed ODM. 
    The leftmost column shows $p^{train}(v|o)$ and the other 4 columns represent the sampled verb distributions at different iterations.
    }
\end{figure}

We show the evolution of label distribution under another four randomly picked objects in Fig.~\ref{fig:memvis}.
It can be observed that the model prefers to sample some frequent class instances at early iterations due to their dominance.
When it comes to later training steps, rare class instances gain more attention with the help of the proposed ODM.
By the end of the first epoch (\textit{i.e.}, ~4.5k iterations), the tail classes under each object is more frequently sampled.  

\subsection{More Qualitative Results}
\begin{figure}[t]
    \centering
    \includegraphics[width=1.0\columnwidth, trim=0 0 0 0, clip]{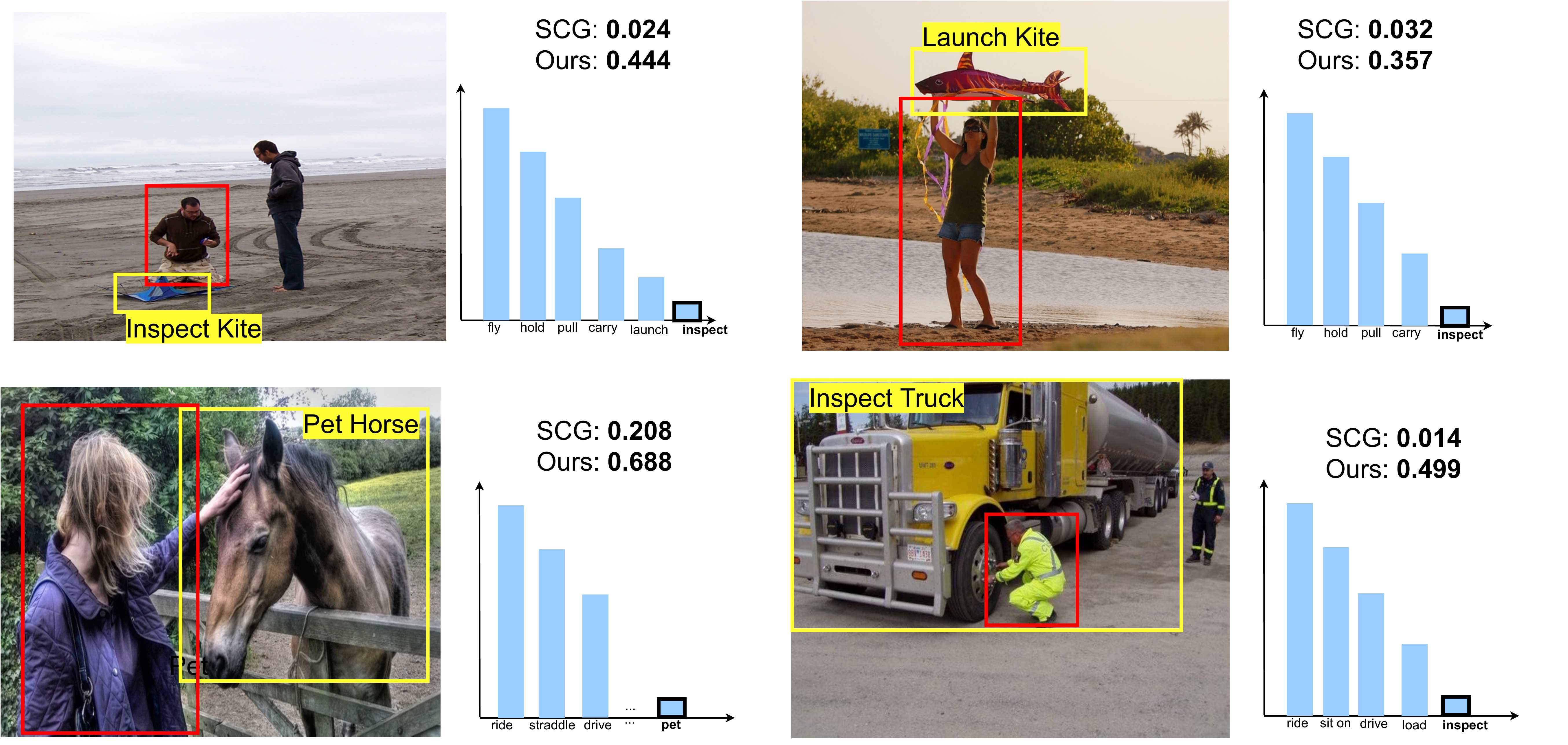}
    \caption{\label{fig:quali_sup}
    Additional qualitative results. 
    Human and object are bounded by red box and yellow box, where the tag indicates the ground truth interaction. 
    For each example, the object-conditional verb distribution on training set $p^{train}(v|o)$ are shown, where the involved verb is bold.
    }
    \vspace{-1em}
\end{figure}

We provide additional qualitative results in Fig.~\ref{fig:quali_sup}.
It can be seen that our method can effectively alleviate the \textit{object bias problem} by reducing false negative errors.

\clearpage
%
%
\bibliographystyle{splncs04}
\bibliography{egbib}